\documentclass[11pt,a4paper]{article}
\usepackage[hyperref]{acl2021}
\usepackage{times}
\usepackage{latexsym}

\usepackage{microtype}

\aclfinalcopy 
\def\aclpaperid{2340} 

\usepackage{amsfonts}       
\usepackage{amsmath}
\usepackage{amssymb}
\usepackage{graphicx}
\usepackage{enumitem}
\usepackage{array}
\usepackage{breakcites}
\usepackage{multirow}
\usepackage{url}

\usepackage{arydshln}
\usepackage{algorithm}          
\usepackage{algpseudocode}      
\usepackage{amsmath}            
\usepackage{amssymb}            

\newcommand{\agggen}{\textsc{AggGen}}  
\newcommand{\argmax}[1]{\underset{#1}{\operatorname{arg}\,\operatorname{max}}\;}

\usepackage{xcolor}
\usepackage[normalem]{ulem}

\def\ODdel#1{\bgroup\markoverwith{\textcolor{cyan!89!yellow!80!black!100}{\rule[0.4ex]{2pt}{3pt}}}\ULon{#1}}
\def\VR#1{{\color{magenta}VR: \it #1}}
\def\VRdel#1{\bgroup\markoverwith{\textcolor{magenta}{\rule[0.5ex]{2pt}{1pt}}}\ULon{#1}}

\def\XXdel#1{\bgroup\markoverwith{\textcolor{green!10!orange!90!}{\rule[0.5ex]{2pt}{1pt}}}\ULon{#1}}

\def\IKdel#1{\bgroup\markoverwith{\textcolor{green!80!black!100}{\rule[0.4ex]{2pt}{3pt}}}\ULon{#1}}

\def\JHdel#1{\bgroup\markoverwith{\textcolor{red!80!black!100}{\rule[0.4ex]{2pt}{3pt}}}\ULon{#1}}

\title{Aggregation before Generation: Factuality Controlled Text Generation}
\title{Planning to Aggregate Language}
\title{Planning to Aggregate in Language Generation}
\title{Planning to Aggregate in Data-to-Text Generation}
\title{Sentence Planning in Data-to-Text Generation}
\title{{\agggen}: Jointly learning sentence planning and realisation in data-to-text generation \VR{Do we need to mention aggregation in the title cf. name of the system?}}
\title{{\agggen}: Ordering and Aggregating while Generating}
\title{{\agggen}: Ordering and Aggregating while Generating}

\author{Xinnuo Xu$^\dagger$, Ond\v{r}ej Du\v{s}ek$^\ddagger$, Verena Rieser$^\dagger$ and Ioannis Konstas$^\dagger$ \\
  $^\dagger$The Interaction Lab, MACS, Heriot-Watt University, Edinburgh, UK \\
  $^\ddagger$Charles University, Faculty of Mathematics and Physics, Prague, Czechia \\
  {\tt xx6, v.t.rieser, i.konstas@hw.ac.uk} \\
  {\tt odusek@ufal.mff.cuni.cz} \\}

\date{}

\begin{document}
\maketitle
\begin{abstract}
We present \agggen{} (pronounced {\em `again'}) a data-to-text model which re-introduces two explicit sentence planning stages into neural data-to-text systems: input ordering and input aggregation.
In contrast to previous work using sentence planning, our model is still end-to-end: \agggen{} performs sentence planning at the same time as generating text by learning latent alignments (via semantic facts) between input representation and target text. 
Experiments on the WebNLG and E2E challenge data show that by using fact-based alignments our approach is more interpretable, expressive, robust to noise,
and easier to control, while retaining the advantages of end-to-end systems in terms of fluency. Our code is available at \url{https://github.com/XinnuoXu/AggGen}.
\end{abstract}


\section{Introduction}\label{sec:Intro}
\begin{figure}[tb]
\includegraphics[width=1\columnwidth]{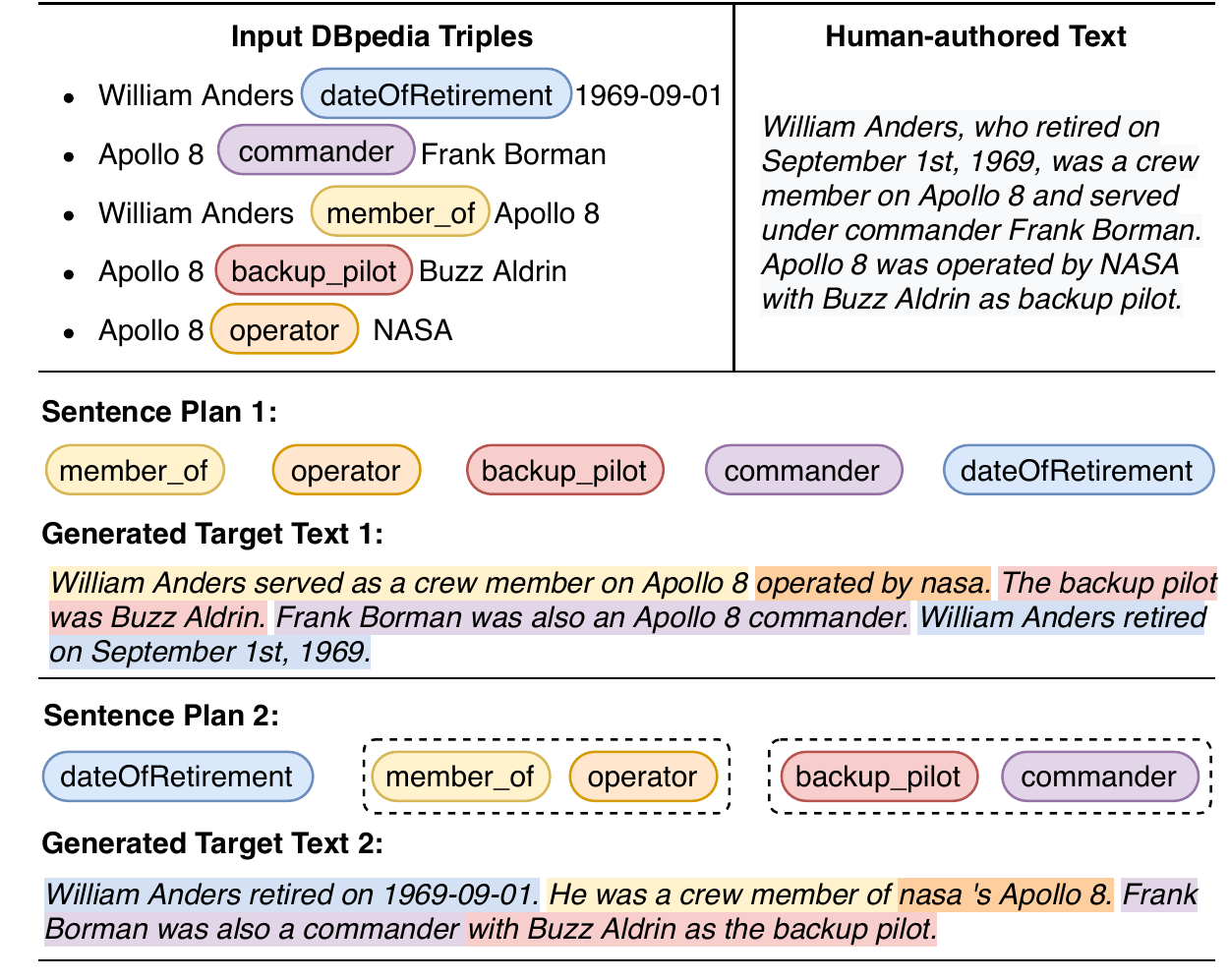}
\caption{Two different sentence plans with their corresponding generated target texts from our model on the WebNLG dataset. Planning and generation is performed jointly. The dashed line denotes aggregation.}
\label{fig:webnlg_ex}
\vspace{-12pt}
\end{figure}

Recent neural data-to-text systems generate text ``end-to-end" (E2E) by learning an implicit mapping between input representations (e.g. RDF triples) and target texts.
While this can lead to increased fluency, E2E methods often produce repetitions, hallucination 
and/or omission of important content for data-to-text \cite{DUSEK:e2e2020} as well as other natural language generation (NLG) tasks \cite{cao_faithful_2018,rohrbach_object_2018}.
Traditional NLG systems, on the other hand, tightly control which content gets generated, as well as its ordering  and aggregation. 
This process is called {\em sentence planning} \cite{reiter2000building,duboue-mckeown-2001-empirically, duboue-mckeown-2002-content,konstas-lapata-2013-inducing,gatt2018survey}. 
Figure~\ref{fig:webnlg_ex} shows two different ways to arrange and combine the representations in the input, resulting in widely different generated target texts.

In this work, we combine advances of both paradigms into a single system by reintroducing sentence planning into neural architectures. 
We call our system  \agggen{} (pronounced {\em `again'}). 
\agggen{} jointly learns to generate and plan at the same time. 
Crucially, our sentence plans are {\em interpretable} latent states using semantic {\em facts}\footnote{Each fact roughly captures “who did what to whom”.} (obtained via Semantic Role Labelling (SRL)) that align the target text with parts of the input representation. In contrast, the plan used in other neural plan-based approaches is usually limited in terms of its interpretability, control, and expressivity.
For example, in \cite{moryossef_step-by-step:_2019,zhao_bridging_2020} the sentence plan is created independently, incurring error propagation; 
\citet{wiseman-etal-2018-learning} use 
latent segmentation 
that limits interpretability; 
\citet{shao2019long} sample from a 
latent variable,
 not allowing for explicit control; 
and \citet{shen2020neural} aggregate multiple input representations which limits expressiveness.

\agggen{} explicitly models the two planning processes (ordering and aggregation), but can directly influence the resulting plan and generated target text, using a separate inference algorithm based on dynamic programming. Crucially, this enables us to directly evaluate and inspect the model's planning and alignment performance by comparing to manually aligned reference texts.

We demonstrate this for two data-to-text generation tasks: the E2E NLG  \cite{novikova2017e2e} and the WebNLG Challenge \cite{gardent-etal-2017-creating}. %
We work with a triple-based semantic representation where a triple consists of a subject, a predicate and an object.\footnote{Note that E2E NLG data and other input semantic representations can be converted into triples, see Section~\ref{subsec:Experiments-dataset}.} 
For instance, in the last triple in \mbox{\autoref{fig:webnlg_ex}}, 
\textit{Apollo 8}, \textit{operator} and \textit{NASA} are the subject, predicate and object respectively.
Our contributions are as follows:

$\bullet$ We present a novel interpretable architecture for jointly learning to plan and generate based on modelling ordering and aggregation by aligning facts in the target text to input representations with an HMM and Transformer encoder-decoder. 

$\bullet$ We show that our method generates output with higher factual correctness than vanilla encoder-decoder models without semantic information. 

$\bullet$ We also introduce an intrinsic evaluation framework for inspecting sentence planning with a rigorous human evaluation procedure to assess factual correctness in terms of alignment, aggregation and ordering performance.


\section{Related Work}\label{sec:Related}

Factual correctness is one of the main issues for data-to-text generation: How to generate text according to the  facts specified in the input triples without adding, deleting or replacing information? 

The prevailing sequence-to-sequence (seq2seq) architectures typically address this issue via reranking 
\cite{wen_stochastic_2015,dusek_sequence--sequence_2016,juraska_deep_2018} or some sophisticated training techniques \cite{nie_simple_2019,kedzie_good_2019,qader_semi-supervised_2019}. 
For applications where structured inputs are present, neural graph encoders \cite{marcheggiani2018deep,rao_tree--sequence_2019,gao_rdf--text_2020} or decoding of explicit graph references \cite{logan_baracks_2019} are applied for higher accuracy.
Recently, large-scale pretraining has achieved SoTA results on WebNLG by fine-tuning T5 \cite{kale-rastogi-2020-text}. 

Several works aim to improve accuracy and controllability  by
dividing the end-to-end architecture into \emph{sentence planning} and \emph{surface realisation}. 
\citet{ferreira_neural_2019} feature a pipeline with multiple planning stages and
\citet{elder2019designing} introduce a symbolic intermediate representation in multi-stage neural generation.
\citet{moryossef_step-by-step:_2019,moryossef_improving_2019}
use pattern matching to approximate the required planning annotation (entity mentions, their order and sentence splits).
\citet{zhao_bridging_2020} use a planning stage in a graph-based model -- the graph is first reordered into a plan; the decoder conditions on both the input graph encoder and the linearized plan.
Similarly, \citet{fan_strategies_2019} use a pipeline approach for story generation via SRL-based sketches.
However, all of these pipeline-based approaches either require additional manual annotation or depend on a parser for the intermediate steps. 


Other works, in contrast, learn planning and realisation jointly.
For example, \citet{su2018natural} introduce a hierarchical decoding model generating different parts of speech at different levels, while filling in slots between previously generated tokens.
\citet{aaai.v33i01.33016908}
include a jointly trained content selection and ordering module that is applied before the main text generation step.
The model is trained by maximizing the log-likelihood of the gold content plan and the gold output text.
\citet{li-rush-2020-posterior} utilize posterior regularization in a structured variational framework to induce which input items are being described 
by each token of the generated text. 
\citet{wiseman-etal-2018-learning} 
aim for better semantic control
by using a Hidden Semi-Markov Model (HSMM) for splitting target sentences into short phrases corresponding to ``templates”, which are then concatenated to produce the outputs. However it trades the controllability for fluency.
Similarly, \citet{shen2020neural} explicitly segment target text into fragment units, while aligning them with their 
corresponding input.
\citet{shao2019long} use a Hierarchical Variational Model to aggregate input items into a sequence of local latent variables and realize sentences conditioned on the aggregations. The aggregation strategy is controlled by sampling from a global latent variable.

In contrast to these previous works, we achieve input ordering and aggregation, input-output alignment and text generation control via {\em interpretable states}, while preserving fluency. 

\section{Joint Planning and Generation}\label{sec:Decoder} 

We jointly learn to generate and plan by aligning {\em facts} in the target text with parts of the input representation.
We model this alignment using a Hidden Markov Model (HMM) that follows a hierarchical structure comprising two sets of latent states, corresponding to ordering and aggregation. 
The model is trained end-to-end 
and all intermediate steps are learned in a unified framework.



\subsection{Model Overview}\label{sec:Decoder-overview}

Let $x=\left \{ x_1,  x_2, \ldots, x_J \right \}$ be a collection of $J$ input triples and $\mathbf{y}$ their natural language description (human written target text). 
We first segment $\mathbf{y}$ into a sequence of $T$ \textit{facts} $\mathbf{y}_{1:T} = \mathbf{y}_1, \mathbf{y}_2, \ldots, \mathbf{y}_T$, where each \textit{fact} roughly captures “who did what to whom” in one event. We follow the approach of \citet{xu2020fact}, where \textit{facts} correspond to predicates and their arguments as identified by SRL (See Appendix~\ref{sec:sup-Segmentation} for more details). For example:
\vspace{-5pt}
\begin{figure}[!h]
\includegraphics[width=1\columnwidth]{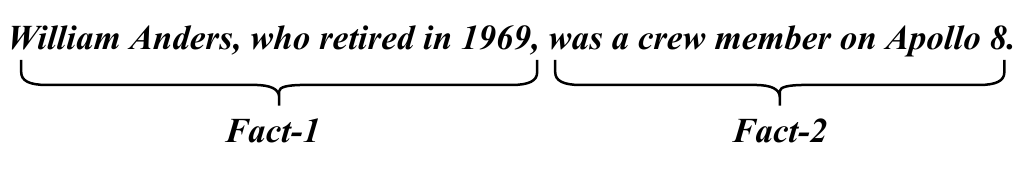}
\vspace{-25pt}
\end{figure}

Each \textit{fact} $\mathbf{y}_t$ consists of a sequence of tokens $y_t^1, y_t^2, \ldots, y_t^{N_t}$.
Unlike the text 
itself, the planning information, i.e.\ input aggregation and ordering, is not directly observable due to the absence of labelled 
datasets. \agggen{} therefore utilises an HMM probabilistic model 
which assumes that there is an underlying hidden process that can be modeled by a first-order Markov chain. At each time step, a latent variable 
(in our case input triples) is responsible for emitting an observed variable (in our case a fact text segment). The HMM specifies a joint distribution on the observations and the latent variables. Here, a latent state $\mathbf{z}_{t}$ emits a \textit{fact} $\mathbf{y}_{t}$, representing the group of input triples that is verbalized in $\mathbf{y}_{t}$. 
We write the joint likelihood as:
\begin{equation*}\label{Eq:1}
\small
\begin{split}
p &\left ( \mathbf{z}_{1:T}, \mathbf{y}_{1:T} \mid x\right ) = p\left ( \mathbf{z}_{1:T} \mid x \right )p\left (\mathbf{y}_{1:T} \mid \mathbf{z}_{1:T}, x \right ) \\
& =\left [ p\left (  \mathbf{z}_1 \mid x \right )\prod_{t=2}^{T} p\left (  \mathbf{z}_t  \mid \mathbf{z}_{t-1}, x \right ) \right ]\left [ \prod_{t=1}^{T} p\left (\mathbf{y}_t \mid \mathbf{z}_t, x  \right ) \right ].
\end{split}
\end{equation*}
i.e., it is a product of the probabilities of each latent state transition (\emph{transition distribution}) and the probability of the observations given their respective latent state (\emph{emission distribution}).

\subsection{Parameterization}\label{sec:Decoder-para}
\begin{figure}[tb]
\includegraphics[width=1.0\columnwidth]{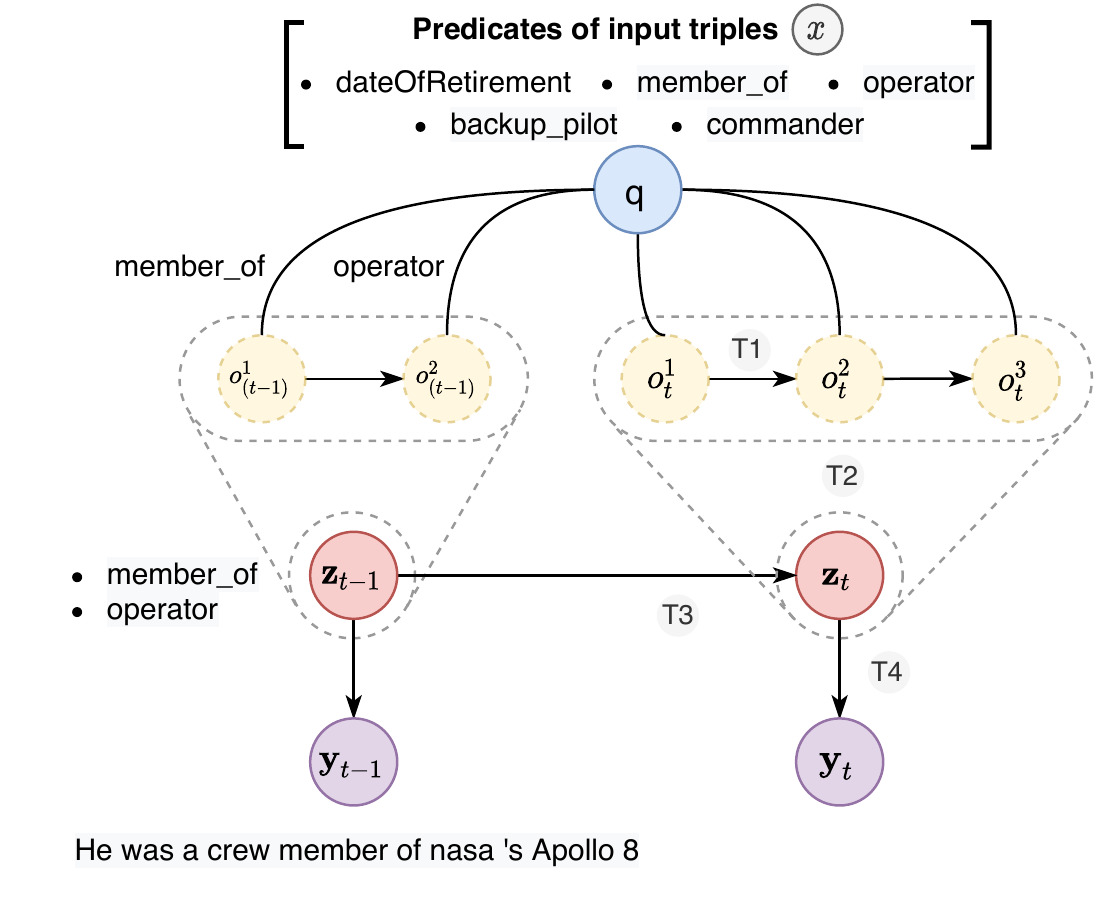}
\centering
\caption{The structure of our model. $\mathbf{z}_t$, $\mathbf{z}_{t-1}$, $\mathbf{y}_t$, and $\mathbf{y}_{t-1}$ represent the basic HMM structure, where $\mathbf{z}_t$, $\mathbf{z}_{t-1}$ are latent states and $\mathbf{y}_t$, $\mathbf{y}_{t-1}$ are observations. Inside the dashed frames is the corresponding structure for each latent state $\mathbf{z}_t$, which is a sequence of latent variables $\mathbf{o}_{t}^{L_t}$ representing the predicates that emit the observation. For example, at time step $t-1$ two input triples (`member\_of' and `operator') are verbalized in the observed \textit{fact} $\mathbf{y}_{t-1}$, whose predicates are represented as latent variables $o_{(t-1)}^1$ and $o_{(t-1)}^2$. 
T1--4 represent transitions introduced in Section~\ref{sec:Decoder-para}. 
}
\label{fig:dag}
\vspace{-12pt}
\end{figure}

\paragraph{Latent State.}\label{par:Decoder-latent}
A latent state $\mathbf{z}_t$ represents the input triples that are verbalized in the observed \textit{fact} $\mathbf{y}_t$. 
It is not guaranteed that one \textit{fact} always verbalizes only one triple (see bottom example in \mbox{\autoref{fig:webnlg_ex}}). 
Thus, we represent state $\mathbf{z}_t$ as a sequence of latent variables $o_{t}^{1}, \ldots, o_{t}^{L_t}$, where $L_t$ is the number of triples 
verbalized in $\mathbf{y}_t$. \autoref{fig:dag} shows the structure of the model.  

Let $o_{t}^l \in \mathbb{Q}=\left \{ 1, \ldots, K \right \}$ be a set of possible latent variables, 
then $K^{L_t}$ is the size of the search space for $\mathbf{z}_t$. 
If $o_{t}^l$ maps to unique triples, the search space becomes intractable for a large value of $K$.
To make the problem tractable, we decrease $K$ by representing triples by their predicate. 
$\mathbb{Q}$ thus stands for the collection of all predicates appearing in the corpus. To reduce the search space for $\mathbf{z}_t$ further, we limit $L_t < \mathbb{L}$, where $\mathbb{L}=3$. \footnote{By aligning the triples to facts using a rule-based aligner (see Section~\ref{sec:Results}), we found that the chance of aggregating more than three triples to a fact is under 0.01\% in the training set of both WebNLG and E2E datasets.}

\paragraph{Transition Distribution.}\label{par:Decoder-trans}
The transition distribution between latent variables (T1 in \autoref{fig:dag}) is a $K \times K$ matrix of probabilities, where each row sums to 1. We define this matrix as
\begin{equation}\label{Eq:3}
 p\left ( o_{t}^l \mid o_{t}^{(l-1)}, x \right ) = \textup{softmax}\left (  AB \odot M\left ( q \right ) \right )
\end{equation}
where $\odot$ denotes the Hadamard product. \mbox{$A\in \mathbb{R}^{K\times m}$} and \mbox{$B \in \mathbb{R}^{m \times K}$} are matrices of 
predicate
embeddings with dimension $m$.
\mbox{$q=\left \{ q_1, q_2, \ldots, q_J \right \}$} is the set of 
predicates of the input triples $x$, and each $q_j \in \mathbb{Q}$ is the predicate of the triple $x_j$. $M\left ( q \right )$ is a ${K\times K}$ masking matrix, where $M_{ij} = 1$ if $i\in q$ and $j\in q$, otherwise $M_{ij} = 0$. 
We apply row-wise softmax over the resulting matrix to obtain probabilities.

The probability of generating the latent state $\mathbf{z}_t$ (T2 in \autoref{fig:dag}) can be written as the joint distribution of the latent variables $o_{t}^1, \ldots, o_{t}^{L_t}$. Assuming a first-order Markov chain, we get:
\vspace{-6pt}
\begin{equation*}\label{Eq:4}
\small
\begin{split}
p\left ( \mathbf{z}_t \mid x \right ) & = p\left ( o_{t}^{0}, o_{t}^{1}, o_{t}^{2}, \ldots, o_{t}^{L_t} \mid x \right )  \\
& = p\left ( o_{t}^{0} \mid x \right )\left [ \prod_{l=1}^{L_{t}} p\left ( o_{t}^{l} \mid o_{t}^{(l-1)}, x \right )  \right ],
\end{split}
\vspace{-6pt}
\end{equation*}
where $o_{t}^{0}$ is a 
marked start-state. 

On top of the generation probability of the latent states $p\left ( \mathbf{z}_t \mid x \right )$ and $p\left ( \mathbf{z}_{t-1} \mid x \right )$, we define the transition distribution between two latent states \mbox{(T3 in \autoref{fig:dag})} as:
\begin{equation*}\label{Eq:5}
\small
\begin{split}
p  \left ( \mathbf{z}_t  \mid \mathbf{z}_{t-1}, x \right ) = &
p\left ( o_{(t-1)}^{0}, \ldots, o_{(t-1)}^{L_{t-1}}  \mid x \right ) \\
 & \cdot p\left ( o_{t}^{1}  \mid o_{(t-1)}^{L_{t-1}}, x \right ) \\ 
 & \cdot p\left ( o_{t}^{0}, \ldots, o_{t}^{L_t}  \mid x \right ),
\end{split}
\end{equation*}
where $o_{(t-1)}^{L_{t-1}}$ denotes the last latent variable in latent state $\mathbf{z}_{t-1}$, while $o_{t1}$ denotes the first latent variable (other than the start-state) in latent state $\mathbf{z}_t$. We use two sets of parameters $\left \{ A_{\textup{in}}, B_{\textup{in}}  \right \}$ and $\left \{ A_{\textup{out}}, B_{\textup{out}} \right \}$ to describe the transition distribution between latent variables within and across latent states, respectively.

\paragraph{Emission Distribution.}\label{par:Decoder-emit} 
The emission distribution $p\left (\mathbf{y}_t \mid \mathbf{z}_t, x  \right )$ (T4 in \autoref{fig:dag}) describes the generation of \textit{fact} $\mathbf{y}_t$ conditioned on latent state $\mathbf{z}_t$ and input triples $x$. We define the probability of generating a \textit{fact} as the product over token-level probabilities, 
\vspace{-12pt}
\begin{equation*}\label{Eq:6}
  \small
  \begin{split}
  p\left (\mathbf{y}_t \mid \mathbf{z}_t, x  \right ) = p(y_t^1 \mid \mathbf{z}_t, x )\prod_{i=2}^{N_t}p(y_t^i \mid  y_t^{1:(i-1)} , \mathbf{z}_t, x ).
  \end{split}
  \vspace{-12pt}
  \end{equation*}
The first and last token of a \textit{fact} are  
marked fact-start and fact-end tokens. We adopt Transformer \cite{vaswani2017attention} as the model's encoder and decoder. 

Each triple is linearized into a list of tokens following the order: subject, predicate, and object. In order to represent individual triples, we insert special \texttt{[SEP]} tokens at the end of each triple. A special \texttt{[CLS]} token is inserted before all input triples, representing the beginning of the entire input. 
An example where the encoder produces a contextual embedding for the tokens of two input triples is shown in \autoref{fig:mask} in Appendix~\ref{sec:sup-emission}.

At time step $t$, the decoder generates \textit{fact} $\mathbf{y}_t$ token-by-token autoregressively, conditioned on both the contextually-encoded input
and the latent state $\mathbf{z}_t$. To guarantee that the generation of $\mathbf{y}_t$ conditions only on the input triples whose predicate is in $\mathbf{z}_t$, we mask out the contextual embeddings of tokens from other unrelated triples for the encoder-decoder attention in all Transformer layers. 

\paragraph{Autoregressive Decoding.}\label{par:Decoder-autoreg} 
Autoregressive Hidden Markov Model (AR-HMM) introduces extra links into HMM to capture long-term correlations between observed variables, i.e., output tokens. Following \newcite{wiseman-etal-2018-learning}, we use AR-HMM for decoding, therefore allowing the interdependence between tokens to generate more fluent and natural text descriptions.
Each token distribution depends on all the previously generated tokens, i.e.,
we define the token-level probabilities as \mbox{$p ( y_t^i \mid  y_{1:(t-1)}^{1:N_t}, y_t^{1:(i-1)} , \mathbf{z}_t, x )$} instead of \mbox{$p ( y_t^i \mid  y_t^{1:(i-1)} , \mathbf{z}_t, x )$}. During training, at each time step $t$, we teacher-force the generation of the fact $\mathbf{y}_t$ by feeding the ground-truth history, $\mathbf{y}_{1:(t-1)}$, to the word-level Transformer decoder. However, since only $\mathbf{y}_t$ depends on the current hidden state $\mathbf{z}_t$, we only calculate the loss over $\mathbf{y}_t$. 

\subsection{Learning}\label{sec:Decoder-learn}
We apply the backward algorithm \cite{rabiner1989tutorial}
to learn the parameters introduced in Section~\ref{sec:Decoder-para}, where we maximize $p(\mathbf{y} \mid x)$, i.e., the marginal likelihood of the observed \textit{facts} $\mathbf{y}$ given input triples $x$, over all the latent states $\mathbf{z}$ and $\mathbf{o}$ on the entire dataset using dynamic programming.
Following \citet{murphy2012machine}, and given that the latent state at time $t$ is $C$, we define a conditional likelihood of future evidence as:
\vspace{-4pt}
\begin{equation}\label{Eq:7}
  \small
  \begin{split}
    \beta_t\left ( C \right ) \triangleq p\left ( \mathbf{y}_{t+1:T} \mid \mathbf{z}_t = C, x \right ),
  \end{split}
\end{equation}
where $C$ denotes a group of predicates that are associated with the emission of $\mathbf{y}$. The size of $C$ ranges from 1 to $\mathbb{L}$ 
and each component is from the collection of predicates $\mathbb{Q}$ (see Section~\ref{sec:Decoder-para}).
Then, the backward recurrences are:
\begin{equation*}\label{Eq:8}
  \small
  \begin{split}
    & \beta_{t-1} \left ( {C}' \right ) = p\left ( \mathbf{y}_{t:T} \mid \mathbf{z}_{t-1} = {C}', x \right ) \\
    &= \sum_{C} \beta_{t}\left ( C \right )p\left (  \mathbf{y}_t \mid  \mathbf{z}_t = C, x \right )p\left (  \mathbf{z}_t = C \mid  \mathbf{z}_{t-1} = {C}', x\right )
  \end{split}
\end{equation*}
with the base case $\beta_T\left ( C \right ) = 1$.
The marginal probability of $\mathbf{y}$ over latent $\mathbf{z}$ is then obtained as $p\left ( \mathbf{y} \mid x \right ) = \sum_C \beta_0\left ( C \right )p\left ( \mathbf{z}_1=C |x \right )$.

In \autoref{Eq:7}, the size of the search space for $C$ is $\sum_{\alpha=1}^{\mathbb{L}}K^\alpha $, where $K = \left | \mathbb{Q} \right |$, i.e., the number of unique predicates appearing in the dataset. The problem can still be intractable due to high $K$, despite the simplifications explained in Section~\ref{sec:Decoder-para} (cf.~predicates).
To tackle this issue and reduce the search space of $C$, we: (1) only explore permutations of $C$ that  include predicates appearing on the input; (2) introduce a heuristic based on the overlap of tokens between a triple and a \textit{fact}---if a certain fact mentions most tokens appearing in the predicate and object of a triple we hard-align it to this triple.\footnote{
    This heuristic is using the rule-based aligner introduced in Section~\ref{sec:Results} with a threshold to rule out alignments in which the triples are not covered over 50\%, since our model emphasises more on precision. Thus, not all triples are aligned to a \textit{fact}.
}
As a result, we discard the permutations that do not include the aligned predicates.

\subsection{Inference}\label{sec:Decoder-infer}
After the joint learning process, the model is able to plan, i.e., order and aggregate the input triples in the most likely way, and then generate a text description following the planning results.
Therefore, the joint prediction of $( \hat{y}, \hat{\mathbf{z}} )$ is defined as:
\vspace{-4pt}
\begin{equation}\label{Eq:9}
  \small
  \begin{split}
    \left ( \hat{y}, \hat{\mathbf{z}} \right ) &= \argmax{\left ( \mathbf{y}', \mathbf{z}' \right ), \mathbf{z}' \in \{ \tilde{\mathbf{z}}^{(i)} \}} p\left ( \mathbf{y}', \mathbf{z}' \mid x \right ) \\
    &= \argmax{\left ( \mathbf{y}', \mathbf{z}' \right ), \mathbf{z}' \in \{ \tilde{\mathbf{z}}^{(i)} \}} p( \mathbf{y}' \mid \mathbf{z}', x) p( \mathbf{z}' \mid x),
  \end{split}
  \vspace{-4pt}
\end{equation}
where $\{ \tilde{\mathbf{z}}^{(i)} \}$ denotes a set of planning results,
$\hat{y}$ is the text description, and
$\hat{\mathbf{z}}$ is the planning result that $\hat{y}$ is generated from. 

\begin{figure}[tb]
  \includegraphics[width=0.8\columnwidth]{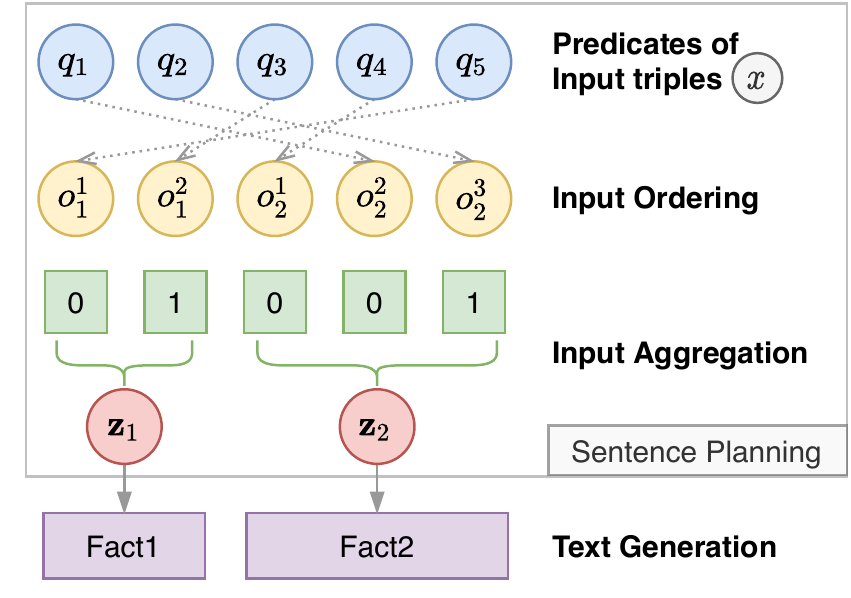}
  \centering
  \caption{The inference process (Section~\ref{sec:Decoder-infer}) 
  }
  \label{fig:inference}
  \vspace{-12pt}
  \end{figure}

The entire inference process (see \autoref{fig:inference}) includes three steps: input ordering, input aggregation, and text generation. The first two steps are responsible for the generation of $\{ \tilde{\mathbf{z}}^{(i)} \}$ together with their probabilities $\{ p( \tilde{\mathbf{z}}^{(i)} \mid x)\}$, while the last step is for the text generation $p( \mathbf{y}' \mid \tilde{\mathbf{z}}^{(i)}, x)$.

\vspace{1ex}
\noindent\textbf{Planning: Input Ordering.}\label{par:Decoder-plan}
The aim is to find the \textit{top-k} most likely orderings of predicates appearing in the input triples.
In order to make the search process more efficient, we apply left-to-right beam-search\footnote{We use beam search since
Viterbi decoding aims at getting $\textbf{z}^* = \arg\max_{\textbf{z}}(\textbf{z}_{1:T} | \textbf{y}_{1:T})$, but $\textbf{y}_{1:T}$ is not available at this stage.} based on the transition distribution introduced in \autoref{Eq:3}.
Specifically, we use a transition distribution between latent variables within latent states, calculated with predicate embeddings $A_{\textup{in}}$ and $B_{\textup{in}}$ (see Section~\ref{sec:Decoder-para}).
To guarantee that the generated sequence does not suffer from omission and duplication of predicates, we constantly update the masking matrix $M(q)$ by removing generated predicates from the set $q$.
The planning process stops when $q$ is empty. 

\noindent\textbf{Planning: Input Aggregation.}\label{par:Decoder-agg}
The goal is to find the top-$n$ 
most likely aggregations for each result of the \textit{Input Ordering} step.
To implement this process efficiently, we introduce a binary state for each predicate in the sequence: 0 indicates ``wait'' and 1 indicates ``emit'' (green squares in \autoref{fig:inference}).
Then we list all possible combinations\footnote{%
  We assume that each fact is comprised of $\mathbb{L}$  triples at most.
  To match this assumption, we discard combinations containing a group that aggregates more than $\mathbb{L}$ predicates.
} of the binary states for the \textit{Input Ordering} result.
For each combination, the aggregation algorithm 
proceeds left-to-right over the predicates and
groups those labelled as ``emit'' with all 
immediately preceding
predicates labelled as ``wait''. 
In turn, we rank all the combinations with the transition distribution introduced in \autoref{Eq:3}. 
In contrast to the \textit{Input Ordering} step,  we use the transition distribution between latent variables across latent states, calculated with predicate embeddings $A_{\textup{out}}$ and $B_{\textup{out}}$.
That is, we do not take into account transitions between two consecutive predicates if they belong to the same group.
Instead, we only consider consecutive predicates across two connected groups, i.e., the \emph{last} predicate of the previous group with the \emph{first} predicate of the following group. 

\noindent\textbf{Text Generation.}\label{par:Decoder-gen}
The final step generates a text description conditioned on the  input triples and the planning result (obtained from the \textit{Input Aggregation} step).
We use beam search and the planning-conditioned generation process described in Section~\ref{sec:Decoder-para} (``Emission Distribution'').

\subsection{Controllability over sentence plans}
\label{sec:control}

While the jointly learnt model is capable of fully automatic generation including the planning step (see Section~\ref{sec:Decoder-infer}), the discrete latent space allows direct access to 
manually control the planning component, which is useful in settings which require increased human supervision and is a unique feature of our architecture.
The plans (latent variables) can be controlled in two ways: (1) hyperparameter. 
Our code offers a hyperparameter that can be tuned to control the level of aggregation: no aggregation, aggregate one, two triples, etc. The model can predict the most likely plan based on the input triples and the hyperparameter and generate a corresponding text description; (2) 
the model can 
directly adopt 
human-written plans, e.g. using the notation {\em [eatType][near customer-rating]}, which translates to: {\em first 
generate 
‘eatType’ as an independent fact and then aggregate the predicates ‘near’ and ‘customer-rating’ in the following fact and generate 
their joint 
description.}

\section{Experiments}\label{sec:Experiments}

\subsection{Datasets}\label{subsec:Experiments-dataset}
We tested our approach on two widely used data-to-text tasks: 
the E2E NLG \cite{novikova2017e2e} and WebNLG\footnote{Since we propose exploring sentence planning and increasing the controllability of the generation model and do not aim for a zero-shot setup, we only focus on the {\bf {\em seen}} category in WebNLG.} \cite{gardent-etal-2017-creating}. 
Compared to E2E, WebNLG is smaller, but contains more predicates and has a larger vocabulary. Statistics 
with examples can be found in Appendix~\ref{sec:sup-data-stats}. We followed the original training-development-test data split for both datasets.

\subsection{Evaluation Metrics}\label{subsec:Experiments-eva}

\noindent\textbf{Generation Evaluation}
focuses on evaluating the generated text with respect to its similarity to human-authored reference sentences.
To compare to previous work, we adopt their associated metrics to evaluate each task. 
The E2E task is evaluated using BLEU \cite{papineni-etal-2002-bleu}, NIST \cite{doddington2002automatic}, ROUGE-L \cite{lin2004rouge}, METEOR \cite{lavie2007meteor}, and CIDEr \cite{vedantam2015cider}. 
WebNLG is evaluated in terms of
BLEU, METEOR, and TER \cite{snover2006study}.

\noindent\textbf{Factual Correctness Evaluation}
tests if the generated text corresponds to the input triples \cite{wen-etal-2015-semantically, reed2018can, dusek2020evaluating}. 
We evaluated 
on the E2E test set using automatic slot error rate (SER),\footnote{
  SER is based on regular expression matching.
  Since only the format of E2E data allows such
  patterns for evaluation,
  we only evaluate factual correctness on the E2E task.
} i.e., an estimation of the occurrence of the input attributes (predicates) and their values in the outputs, implemented by \citet{dusek2020evaluating}. 
SER counts predicates that were added, missed or replaced with a wrong object.

\noindent\textbf{Intrinsic Planning Evaluation} examines planning performance in \autoref{sec:intrinsic}.

\subsection{Baseline model and Training Details}\label{subsec:Experiments-detail}
 \vspace{-0.5ex}
To evaluate the contributions of the planning component, we choose the vanilla Transformer model \cite{vaswani2017attention} as our baseline, trained on pairs of linearized input triples and target texts. In addition, we choose two types of previous works for comparison: (1) best-performing models reported on the WebNLG~2017 (seen) and E2E dataset, 
i.e.\ T5 \cite{kale-rastogi-2020-text}, PlanEnc \cite{zhao_bridging_2020}, ADAPT \cite{gardent_webnlg_2017},
and TGen \cite{dusek-jurcicek-2016-sequence}; (2) models with explicit planning, i.e.\ TILB-PIPE \cite{gardent_webnlg_2017}, NTemp+AR \cite{wiseman-etal-2018-learning} and \citet{shen2020neural}. 

To make our HMM-based approach converge faster, we initialized its encoder and decoder with the baseline model parameters and fine-tuned them during training of the transition distributions. 
Encoder and decoder parameters were chosen based on validation results of the baseline model for each task (see Appendix~\ref{sec:sup-hyperparams} for details).

\vspace{-2pt}
\section{Experiment Results}\label{sec:Results}
\vspace{-2pt}
\subsection{Generation Evaluation Results}

\begin{table}[t]
  \small
  \centering
  \begin{tabular}{l|ccc}\hline
    {\bf Model} & {\bf BLEU} & {\bf TER}  & {\bf METEOR} \\ \hline\hline
    {T5}$^\blacklozenge$ & {\bf 64.70} & --- & {\bf 0.46} \\
    {PlanEnc}$^\blacklozenge$ & 64.42 & \bf 0.33 & 0.45 \\
    {ADAPT}$^\blacklozenge$ & 60.59 & 0.37 & 0.44 \\
    TILB-PIPE$^\blacklozenge$ & 44.34 & 0.48 & 0.38 \\
    Transformer & 58.47 & 0.37 & 0.42 \\
    {\agggen} & 58.74 & 0.40 & 0.43 \\
    {\agggen}$_{-\textup{OD}}$ & 55.30 & 0.44 & 0.43 \\
    {\agggen}$_{-\textup{AG}}$ & 52.17 & 0.50 & 0.44 \\ \hline
  \end{tabular}
  \caption{\emph{Generation Evaluation Results} on the WebNLG (seen). The models labelled with $^\blacklozenge$ are from previous work. The rest are our implementations.}
  \label{table:webnlg_res}
  \vspace{-12pt}
\end{table}

\begin{table*}[t]
  \small
  \centering
  \begin{tabular}{l|ccccc|cccc}\hline
  {\bf Model} & {\bf BLEU} & {\bf NIST}  & {\bf MET} & {\bf R-L} & {\bf CIDer} & {\bf Add}  & {\bf Miss} & {\bf Wrong} & {\bf SER} \\ \hline\hline
  TGen$^\blacklozenge$ & 66.41 & 8.5565 & 45.07 & 69.17 & 2.2253 & {\bf00.14} & 04.11 & {\bf00.03} & 04.27 \\ 
  NTemp+AR$^\blacklozenge$ & 59.80 & 7.5600 & 38.75 & 65.01 & 1.9500 & --- & --- & --- & --- \\
  {\citet{shen2020neural}}$^\blacklozenge$ & 65.10 & --- & {\bf45.50} & 68.20 & {\bf2.2410} & --- & --- & --- & --- \\
  Transformer & {\bf68.23} & {\bf8.6765} & 44.31 & {\b69.88} & 2.2153 & 00.30 & 04.67 & 00.20 & 05.16 \\
  \agggen & 64.14 & 8.3509 & 45.13 & 66.62 & 2.1953 & 00.32 & 01.66 & 00.71 & {\bf02.70} \\ 
  {\agggen}$_{-\textup{OD}}$ & 58.90 & 7.9100 & 43.21 & 62.12 & 1.9656 & 01.65 & 02.99 & 03.01 & 07.65 \\
  {\agggen}$_{-\textup{AG}}$ & 44.00 & 6.0890 & 43.75 & 58.24 & 0.8202 & 08.74 & {\bf00.45} & 00.92 & 10.11 \\
  \hline

  \end{tabular}
  \caption{Evaluation of \emph{Generation} (middle) and \emph{Factual correctness} (right) \textbf{\textit{trained/tested}} on the \textbf{\textit{original}} E2E data (Section~\ref{sec:Results} for metrics description).
  Models with $^\blacklozenge$ are from previous work, the rest are our implementations.}
  \label{table:e2e_res}
  \vspace{-10pt}
  \end{table*}

\autoref{table:webnlg_res} shows the generation 
results on the WebNLG seen category \cite{gardent_webnlg_2017}.
Our model outperforms TILB-PIPE and Transformer, 
but performs worse than T5, PlanEnc and ADAPT. However, unlike these three models, our approach does not rely on large-scale pretraining, extra annotation, or heavy pre-processing using external resources.
\autoref{table:e2e_res} shows the results when training and testing on the original E2E set.
{\agggen} outperforms NTemp+AR and is comparable with \citet{shen2020neural}, but performs slightly worse than both seq2seq models in terms of word-overlap metrics. 

However, the results in \autoref{table:e2e_clean_res} demonstrate that our model does outperform the baselines on most surface metrics if trained on the noisy original E2E training set and tested on clean E2E data \cite{dusek-etal-2019-semantic}. 
This suggests that the previous performance drop was due to 
text references in the original dataset that did not verbalize all triples or added information not present in the triples that may have down-voted the fact-correct generations.%
\footnote{We also trained and tested models on the cleaned E2E data. The full results (including the factual correctness evaluation) are shown in \autoref{table:e2e_res-sup} in Appendix~\ref{sec:sup-e2e_res}: there is a similar trend as in results in \autoref{table:e2e_clean_res}, compared to Transformer.}
This also shows that {\agggen} produces correct outputs even when trained on a noisy dataset. Since constructing high-quality data-to-text training sets is expensive and labor-intensive, this robustness towards noise is important.

\begin{table}[t]
  \small
  \centering
  \begin{tabular}{l|p{0.6cm}p{0.6cm}p{0.6cm}p{0.6cm}p{0.6cm}}\hline

  {\bf Model} & {\bf BLEU} & {\bf NIST} & {\bf MET} & {\bf R-L} & {\bf CIDer} \\ \hline\hline
  TGen$^\blacklozenge$ & 39.23 & 6.022 & 36.97 & {\bf55.52} & 1.762 \\ 
  Transformer & 38.57 & 5.756 & 35.92 & 55.45 & 1.668 \\
  {\agggen} & {\bf41.06} & {\bf6.207} & {\bf37.91} & 55.13 & {\bf1.844} \\ 
  {\agggen}$_{-\textup{OD}}$ & 38.24 & 5.951 & 36.56 & 51.53 & 1.653 \\
  {\agggen}$_{-\textup{AG}}$ & 30.44 & 4.636 & 37.99 & 49.94 & 0.936 \\ \hline

  \end{tabular}
  \caption{Evaluation of \emph{Generation} \textbf{\textit{trained}} on the \textbf{\textit{original}} E2E data, while \textbf{\textit{tested}} on the \textbf{\textit{cleaned}} E2E data. Note that, the clean test set has more diverse MRs and fewer references per MR, which leads to lower scores – see also the paper introducing the cleaned E2E data (Table 2 and 3 in \citet{dusek-etal-2019-semantic}).
  }
  \label{table:e2e_clean_res}
  \vspace{-12pt}
\end{table}

\subsection{Factual Correctness Results}
The results for factual correctness evaluated using SER on the original E2E test set are shown in \autoref{table:e2e_res}. 
The SER of {\agggen} is the best among all models. 
Especially, the high ``Miss'' scores for TGen and Transformer demonstrate the high chance of information omission in vanilla seq2seq-based generators. In contrast, {\agggen} shows much better coverage over the input triples while keeping a low level of hallucination (low ``Add'' and ``Wrong'' scores).


\subsection{Ablation variants}

To explore the effect of input planning on text generation, we introduced two model variants: 
{\agggen}$_{-\textup{OD}}$, where we replaced the \textit{Input Ordering} with randomly shuffling the input triples before input aggregation, and {\agggen}$_{-\textup{AG}}$, where the \textit{Input Ordering} result was passed directly to the text generation and the text decoder generated a fact for each input triple individually.

The generation evaluation results on both datasets (\autoref{table:webnlg_res} and \autoref{table:e2e_res}) show that {\agggen} outperforms {\agggen}$_{-\textup{OD}}$ and  {\agggen}$_{-\textup{AG}}$ substantially, 
which means both \textit{Input Ordering} and \textit{Input Aggregation} are critical. %
\autoref{table:e2e_res} shows that the factual correctness results for the ablative variants are much worse than full {\agggen},
indicating that planning is essential for factual correctness.
An exception is the lower number of missed slots in {\agggen}$_{-\textup{AG}}$. This is expected since 
 {\agggen}$_{-\textup{AG}}$
generates a textual fact for each triple individually,
which decreases the possibility of omissions at the cost of much lower fluency. This strategy also leads to a steep increase in added information.

Additionally, {\agggen}$_{-\textup{AG}}$ performs even worse on the E2E dataset than on the WebNLG set. This result is also expected, since input aggregation is more pronounced in the E2E dataset with a higher number of facts and input triples per sentence
(cf.~Appendix~\ref{sec:sup-data-stats}).

\vspace{-3pt}
\subsection{Qualitative Error Analysis}
\vspace{-3pt}

\begin{figure}[tb]
  \includegraphics[width=1.0\columnwidth]{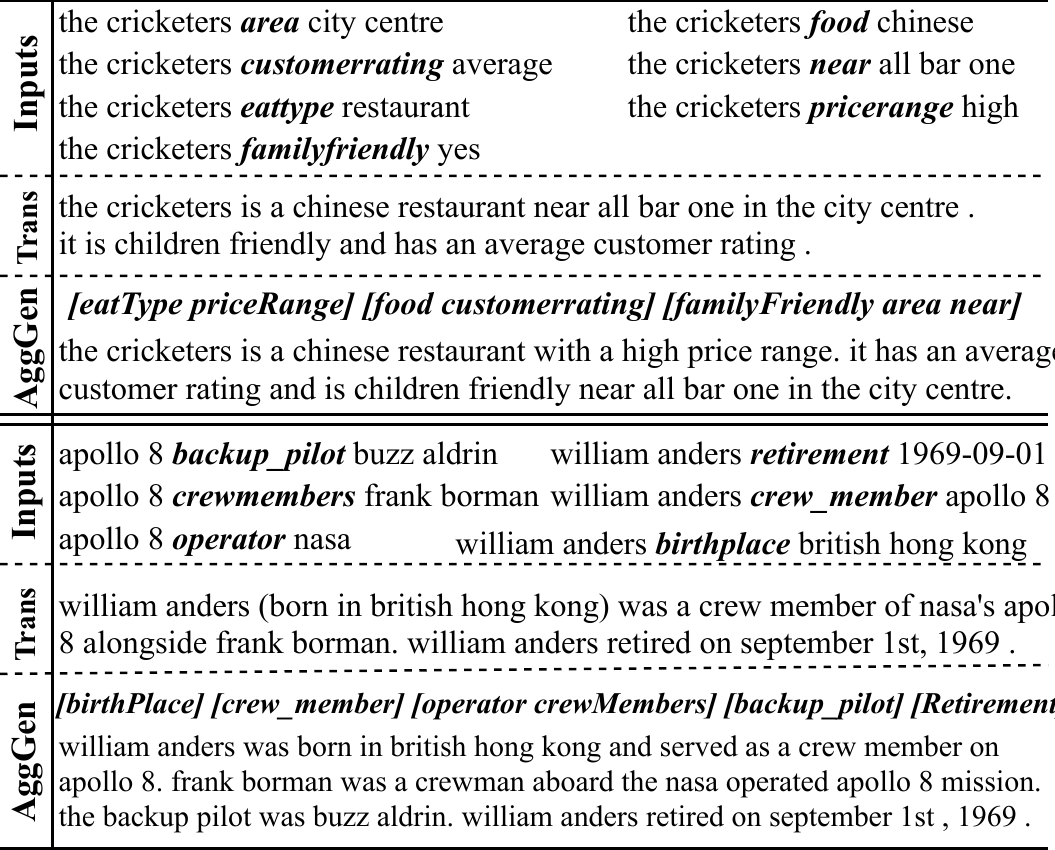}
  \centering
  \caption{Examples of input and system-generated target text for E2E (top) and WebNLG (bottom). The sequences in square brackets are the sentence plans.}
  \label{fig:show}
  \vspace{-12pt}
\end{figure}

We manually examined a sample of 100 outputs (50 from each dataset) with respect to their factual correctness and fluency. 
For factual correctness, we follow the definition of SER and check whether there are hallucinations, substitutions or omissions in generated texts.
For fluency, we check whether the generated texts suffer from grammar mistakes, redundancy, or contain unfinished sentences. 
\autoref{fig:show} shows two examples of generated texts from Transformer and {\agggen} (more examples, including target texts generated by {\agggen}$_{-\textup{OD}}$ and {\agggen}$_{-\textup{AG}}$, are shown in \autoref{table:e2e_res_extra} and \mbox{\autoref{table:webnlg_res_extra}} in Appendix~\ref{sec:sup-example}).
We observe that, in general, the seq2seq Transformer model tends to compress more triples into one fluent fact, whereas {\agggen} aggregates triples in more but smaller groups, and generates a shorter/simpler fact for each group.
Therefore, the texts generated by Transformer are more  compressed, while {\agggen}'s generations are longer with more sentences. 
However, the planning ensures that all input triples will still be mentioned.
Thus, {\agggen} generates texts with higher factual correctness without trading off fluency\footnote{
    The number of fluent generations for Transformer and {\agggen} among the examined 100 examples are 96 and 95 respectively. The numbers for {\agggen}$_{-\textup{OD}}$ and {\agggen}$_{-\textup{AG}}$ are 86 and 74, which indicates that both \textit{Input Ordering} and \textit{Input Aggregation} are critical for generating fluent texts.}. 

\vspace{-3pt}
\section{Intrinsic Evaluation of Planning}
\vspace{-3pt}
\label{sec:intrinsic}
We now directly inspect the performance of the planning component by taking advantage of the readability of SRL-aligned facts. In particular, we investigate:
(1) Sentence planning performance. We study the agreement between model's planning and reference 
planning for the same set of input triples; (2) Alignment performance -- we use \agggen{} 
as an aligner and examine its ability to align segmented facts to the corresponding input triples. 
Since both studies require ground-truth triple-to-fact alignments, which are not 
part of the WebNLG and E2E data, we first introduce a human annotation process in Section~\ref{sec:human-alignments}.

\subsection{Human-annotated Alignments} \label{sec:human-alignments}

We asked crowd workers on Amazon Mechanical Turk to align input triples to their fact-based text snippets
to derive a ``\textbf{\textit{reference plan}}'' for each target text.
\footnote{The evaluation requires human annotations, since anchor-based automatic alignments are not accurate enough (86\%) for the referred plan annotation. See \autoref{table:viterbi} (“RB”) for details.} 
Each worker was given a set of input triples and a corresponding reference text description, segmented into a sequence of facts.
The workers were then asked to select the triples that are verbalised in each fact.\footnote{
  The annotation guidelines and an example annotation task are shown in 
  \autoref{fig:HIT} in Appendix~\ref{sec:sup-annotation}.
}
We sampled 100 inputs from the WebNLG\footnote{
  We chose WebNLG over E2E for its domain and predicate diversity. 
} test set for annotation.
Each input was paired with \textbf{\textit{three reference target texts}}
from WebNLG.
To guarantee the correctness of the annotation, three different workers annotated each input-reference pair.
We only consider the alignments where all three annotators agree.
Using Fleiss Kappa \cite{fleiss1971measuring} over the facts aligned by each judge to each triple, we obtained an average agreement of 0.767 for the 300 input-reference pairs, 
which is considered high agreement.

\subsection{Study of Sentence Planning}

We now check the agreement between the model-generated and 
reference plans based on
the \mbox{top-1} \textit{Input Aggregation} result (see Section~\ref{sec:Decoder-infer}). 
We introduce two metrics: 

$\bullet$ \textit{Normalized Mutual Information (NMI)} \cite{strehl2002cluster}  to evaluate aggregation. We represent
each plan as a set of clusters of triples, where a cluster contains the triples sharing the same fact verbalization. Using NMI we measure mutual information between two clusters, normalized into the 0-1 range, where 0 and 1 denote no mutual information and perfect correlation, respectively.


$\bullet$ \textit{Kendall's tau ($\tau$)} \cite{kendall1945treatment} is a ranking based measure which we use to evaluate both ordering and aggregation. We represent
each plan as a ranking of the input triples, where the rank of each triple is the position of its associated fact verbalization in the target text. $\tau$ measures rank correlation, ranging from -1 (strong disagreement) to 1 (strong agreement).


\begin{table}[t]
  \small
  \centering
  \begin{tabular}{l|cccc}\hline
      & {\bf $\textup{NMI}_\textup{max}$} & {\bf $\textup{NMI}_\textup{avg}$} & {\bf $\textup{K-tau}_\textup{max}$}  & {\bf $\textup{K-tau}_\textup{avg}$} \\ \hline\hline
     {\bf Human} & 0.9340 & 0.7587 & 0.8415 & 0.2488 \\ \hline
     {\bf \agggen} & 0.7101 & 0.6247 & 0.6416 & 0.2064\\ \hline
  \end{tabular}
  \caption{\emph{Planning Evaluation Results}. NMI and \mbox{K-tau} calculated between human-written references (bottom), and between references and our system {\agggen} (top).} 
  \label{table:nmi_tau} 
  \vspace{-8pt}
\end{table}

In the 
crowdsourced annotation (Section~\ref{sec:human-alignments}), each set of input triples contains three reference texts with annotated plans.
We fist evaluate the correspondence among these three reference plans
by calculating NMI and $\tau$ between one  
plan and the remaining two.
In the top row of \autoref{table:nmi_tau}, 
the high average and maximum NMI indicate that the reference texts' authors
tend to aggregate input triples in similar ways.
On the other hand, the low average $\tau$ shows that they are likely to order the aggregated groups differently.
Then, for each set of input triples, we measure NMI and $\tau$ of the top-1 \textit{Input Aggregation} result (model's plan) against each of the corresponding reference plans and compute average and maximum values (bottom row in \autoref{table:nmi_tau}). Compared to the strong agreement among reference plans on the input aggregation, the agreement between model's and reference plans is slightly weaker. 
Our model has slightly lower agreement on aggregation (NMI), 
but if we consider aggregation and ordering jointly ($\tau$), the agreement between our model's plans and reference plans is comparable to the agreement among reference plans. 

\subsection{Study of Alignment} 
\begin{table}[t]
  \small
  \centering
  \begin{tabular}{l|crc}\hline
     & Precision & Recall & F1 \\ \hline\hline
    {\bf RB} (\%) & 86.20 & 100.00 & 92.59 \\ 
    {\bf Vtb} (\%) & 89.73 & 84.16 & 86.85 \\ \hline
  \end{tabular}
  \caption{\emph{Alignment Evaluation Results}. Alignment accuracy for the Viterbi algorithm (Vtb) and the rule-based aligner (RB).}
  \label{table:viterbi}
  \vspace{-12pt}
\end{table}

In this study, we use the HMM model as an aligner and assess its ability to align
input triples with their fact verbalizations on the human-annotated set. 
Given the sequence of observed variables, a trained HMM-based model is able to find the most likely sequence of hidden states $\textbf{z}^* = \underset{\textbf{z}}{\arg\max}(\textbf{z}_{1:T} | \textbf{y}_{1:T})$ using Viterbi decoding. Similarly, given a set of input triples and a factoid segmented text, we use  Viterbi with our model to align each fact with the corresponding input triple(s).
We then evaluate the accuracy of the model-produced alignments against the crowdsourced alignments. 

The alignment evaluation results are shown in \autoref{table:viterbi}.
We compare the Viterbi (Vtb) alignments with the ones calculated by a rule-based aligner (RB) that aligns each triple to the fact with the greatest word overlap.
The precision of the Viterbi aligner is higher than the rule-based aligner.
However, the Viterbi aligner tends to miss triples, which leads to a lower recall.
Since HMMs are locally optimal, the model cannot guarantee to annotate input triples once and only once.

\vspace{-6pt}
\section{Conclusion and Future Work}\label{sec:Future}
\vspace{-6pt}

We show that explicit sentence planning, i.e., input ordering and aggregation, 
helps substantially 
to produce output which is both semantically correct as well as naturally sounding. Crucially, this also enables us to directly evaluate and inspect both the model's planning and alignment performance by comparing to manually aligned reference texts. Our system outperforms vanilla seq2seq models when considering semantic accuracy and word-overlap based metrics. Experiment results also show that {\agggen} is robust to noisy training data. We plan to extend this work in three directions:

\noindent\textbf{Other Generation Models.} We plan to plug other text generators, e.g.\ pre-training based approaches \cite{lewis-etal-2020-bart,kale-rastogi-2020-text} 
, into \agggen{} to enhance their interpretability and controllability via sentence planning and generation.

\noindent\textbf{Zero/Few-shot scenarios.} \citet{kale-rastogi-2020-template}'s work on low-resource NLG uses a pre-trained language model with a schema-guided representation and hand-written templates to guide the representation in unseen domains and slots. These techniques can be plugged into \agggen{}, which allows us to examine the effectiveness of the explicit sentence planning in zero/few-shot scenarios.

\noindent\textbf{Including Content Selection.} In this work, we concentrate on the problem of faithful surface realization based on E2E and WebNLG data, which both operate under the assumption that all input predicates have to be realized in the output. In contrast, more challenging tasks such as RotoWire \cite{wiseman-etal-2017-challenges}, include content selection before sentence planning. 
In the future, we plan to include a content selection step to further extend \agggen{}'s usability.

\vspace{-8pt}
\section*{Acknowledgments}
\vspace{-8pt}
This research received funding from the EPSRC project AISec (EP/T026952/1), Charles University project PRIMUS/19/SCI/10, a Royal Society research grant (RGS/R1/201482), a Carnegie Trust incentive grant (RIG009861), and an Apple NLU Research Grant to support research at Heriot-Watt University and Charles University. We thank Alessandro Suglia, Jind\v{r}ich Helcl, and Henrique Ferrolho for suggestions on the draft. We thank the anonymous reviewers for their helpful comments.

\bibliography{acl2021}
\bibliographystyle{acl_natbib}

\clearpage
\onecolumn
\appendix

\section{Examples of input and system-generated target text}
\label{sec:sup-example}

\begin{table}[!tbh]
  \small
  \centering
  \caption{Examples of input and system-generated target text for E2E NLG dataset.}
  \begin{tabular}{l|p{13.5cm}}\hline \hline
    \multirow{3}{*}{\textbf{Input}} & the cricketers \textbf{\textit{[customerrating]}} high \\
    & the cricketers \textbf{\textit{[eattype]}} restaurant \\
    & the cricketers \textbf{\textit{[near]}} express by holiday inn \\ \hdashline
    \textbf{Transf} & \textbf{Predicted Text:} the cricketers is a highly rated restaurant near express by holiday inn. \\ \hdashline
    \multirow{2}{*}{\textbf{\agggen}} & \textbf{Sentence Planing: \textit{[eatType] [customerrating near]}} \\
    & \textbf{Predicted Text:} the cricketers is a highly rated restaurant located near express by holiday inn. \\ \hdashline
    \multirow{2}{*}{\textbf{{\agggen}$_{-\textup{AG}}$}} & \textbf{Sentence Planing: \textit{[eatType] [near] [customerrating]}} \\
    & \textbf{Predicted Text:} the cricketers is a restaurant located near express by holiday inn . it has a high customer rating. \\ \hdashline
    \multirow{2}{*}{\textbf{{\agggen}$_{-\textup{OD}}$}} & \textbf{Sentence Planing: \textit{[eatType near] [customerrating]}} \\
    & \textbf{Predicted Text:} the cricketers is a restaurant near express by holiday inn . it has a high customer rating. \\ \hline \hline
    
    \multirow{4}{*}{\textbf{Input}} & blue spice \textbf{\textit{[eattype]}} pub \\
    & blue spice \textbf{\textit{[familyfriendly]}} yes \\
    & blue spice \textbf{\textit{[food]}} english \\
    & blue spice \textbf{\textit{[near]}} rainbow vegetarian café \\ \hdashline
    \textbf{Transf} & \textbf{Predicted Text:} the blue spice is a family friendly pub near the rainbow vegetarian café. it serves english food in the higher price range. \\ \hdashline
    \multirow{2}{*}{\textbf{\agggen}} & \textbf{Sentence Planing: \textit{[eatType familyFriendly] [food near]}} \\
    & \textbf{Predicted Text:} blue spice is a family friendly pub serving english food near the rainbow vegetarian café. \\ \hdashline
    \multirow{2}{*}{\textbf{{\agggen}$_{-\textup{AG}}$}} & \textbf{Sentence Planing: \textit{[eatType] [familyFriendly] [food] [near]}} \\
    & \textbf{Predicted Text:} blue spice is a mid - priced pub that is family friendly and serves british food . it is located near the rainbow vegetarian café. \\ \hdashline
    \multirow{2}{*}{\textbf{{\agggen}$_{-\textup{OD}}$}} & \textbf{Sentence Planing: \textit{[near familyFriendly eatType] [food]}} \\
    & \textbf{Predicted Text:} blue spice is a family friendly pub near the rainbow vegetarian café . it serves english food. \\ \hline \hline

    \multirow{6}{*}{\textbf{Input}} & the cricketers \textbf{\textit{[area]}} riverside \\
    & the cricketers \textbf{\textit{[customerrating]}} 5 out of 5 \\
    & the cricketers \textbf{\textit{[eattype]}} restaurant \\
    & the cricketers \textbf{\textit{[familyfriendly]}} yes \\ 
    & the cricketers \textbf{\textit{[food]}} chinese \\
    & the cricketers \textbf{\textit{[near]}} all bar one \\ \hdashline
    \textbf{Transf} & \textbf{Predicted Text:} the cricketers is a family friendly chinese restaurant near all bar one in the riverside area . it has a customer rating of 5 out of 5. \\ \hdashline
    \multirow{2}{*}{\textbf{\agggen}} & \textbf{Sentence Planing: \textit{[eatType familyFriendly] [food area near] [customerrating]}} \\
    & \textbf{Predicted Text:} the cricketers is a family friendly chinese restaurant located in the riverside area near all bar one . it has a customer rating of 5 out of 5. \\ \hdashline
    \multirow{2}{*}{\textbf{{\agggen}$_{-\textup{AG}}$}} & \textbf{Sentence Planing: \textit{[eatType] [food] [area] [customerrating] [familyFriendly] [near]}} \\
    & \textbf{Predicted Text:} the cricketers is a restaurant that serves chinese food. it is located in the riverside area and has a customer rating of 5 out of 5 and is children friendly. the coffee shop is near all bar one. \\ \hdashline
    \multirow{2}{*}{\textbf{{\agggen}$_{-\textup{OD}}$}} & \textbf{Sentence Planing: \textit{[customerrating] [familyFriendly eatType] [food near area]}} \\
    & \textbf{Predicted Text:} the cricketers is a 5 out of 5 rated , family friendly chinese restaurant located in riverside near all bar one. \\ \hline \hline

    \multirow{7}{*}{\textbf{Input}} & the cricketers \textbf{\textit{[area]}} city centre \\
    & the cricketers \textbf{\textit{[customerrating]}} average \\
    & the cricketers \textbf{\textit{[eattype]}} restaurant \\
    & the cricketers \textbf{\textit{[familyfriendly]}} yes \\
    & the cricketers \textbf{\textit{[food]}} chinese \\
    & the cricketers \textbf{\textit{[near]}} all bar one \\
    & the cricketers \textbf{\textit{[pricerange]}} high \\\hdashline
    \textbf{Transf} & \textbf{Predicted Text:} the cricketers is a chinese restaurant near all bar one in the city centre . it is children friendly and has an average customer rating. \\ \hdashline
    \multirow{2}{*}{\textbf{\agggen}} & \textbf{Sentence Planing: \textit{[eatType priceRange] [food customerrating] [familyFriendly area near]}} \\
    & \textbf{Predicted Text:} the cricketers is a chinese restaurant with a high price range . it has an average customer rating and is children friendly near all bar one in the city centre. \\ \hdashline
    \multirow{2}{*}{\textbf{{\agggen}$_{-\textup{AG}}$}} & \textbf{Sentence Planing: \textit{[eatType] [priceRange] [food] [customerrating] [familyFriendly] [area] [near]}} \\
    & \textbf{Predicted Text:} the cricketers is a chinese restaurant that is children friendly, has a high price range, a customer rating of 3 out of 5, is located near the portland arms and is in the city centre. it has an average customer rating. is also a children-friendly . you can find it is called the cricketers. \\ \hdashline
    \multirow{2}{*}{\textbf{{\agggen}$_{-\textup{OD}}$}} & \textbf{Sentence Planing: \textit{[area] [familyFriendly priceRange] [eatType food near] [customerrating]}} \\
    & \textbf{Predicted Text:} located in the city centre, the cricketers is a child friendly chinese restaurant with a high price range. it is near all bar one. its customer rating is average. \\ \hline \hline

  \end{tabular}
  \label{table:e2e_res_extra}
  \vspace{-12pt}
\end{table}

\begin{table}[!tbh]
  \small
  \centering
  \caption{Examples of input and system-generated target text for WebNLG dataset.}
  \begin{tabular}{l|p{13.5cm}}\hline \hline

    \multirow{5}{*}{\textbf{Input}} & 11th mississippi infantry monument \textbf{\textit{[category]}} contributing property \\
    & 11th mississippi infantry monument \textbf{\textit{[country]}} united states \\
    & 11th mississippi infantry monument \textbf{\textit{[established]}} 2000 \\ 
    & 11th mississippi infantry monument \textbf{\textit{[location]}} seminary ridge \\
    & 11th mississippi infantry monument \textbf{\textit{[municipality]}} gettysburg, pennsylvania \\ \hdashline
    \textbf{Transf} & \textbf{Predicted Text:} the 11th mississippi infantry monument is located at seminary ridge, gettysburg, pennsylvania, united states. it was established in 2000 and falls under the category of contributing property. \\ \hdashline
    \multirow{2}{*}{\textbf{\agggen}} & \textbf{Sentence Planing: \textit{[location country established] [category] [municipality]}} \\
    & \textbf{Predicted Text:} the 11th mississippi infantry monument, established in 2000, is located at seminary ridge, gettysburg, pennsylvania, united states. it is categorized as a contributing property and is in the municipality of gettysburg pennsylvania. \\ \hdashline
    \multirow{2}{*}{\textbf{{\agggen}$_{-\textup{AG}}$}} & \textbf{Sentence Planing: \textit{[country] [established] [category] [municipality] [location]}} \\
    & \textbf{Predicted Text:} the 11th mississippi infantry monument is located in gettysburg, pennsylvania, united states. the monument was established in 2000 and falls under the category of contributing property. it can be found in the municipality of gettysburg in pennsylvania and is in the seminary ridge. \\ \hdashline
    \multirow{2}{*}{\textbf{{\agggen}$_{-\textup{OD}}$}} & \textbf{Sentence Planing: \textit{[municipality] [category] [country established] [location]}} \\
    & \textbf{Predicted Text:} the 11th mississippi infantry monument is in the municipality of gettysburg, pennsylvania. it is categorized as a contributing property. the monument was established in the year 2000 in the united states and is located at seminary ridge. \\ \hline \hline

    \multirow{6}{*}{\textbf{Input}} & apollo 8 \textbf{\textit{[backup\_pilot]}} buzz aldrin \\
    & apollo 8 \textbf{\textit{[crewmembers]}} frank borman \\
    & apollo 8 \textbf{\textit{[operator]}} nasa \\
    & william anders \textbf{\textit{[birthplace]}} british hong kong \\
    & william anders \textbf{\textit{[dateofretirement]}} 1969-09-01 \\
    & william anders \textbf{\textit{[was\_a\_crew\_member\_of]}} apollo 8 \\ \hdashline
    \textbf{Transf} & \textbf{Predicted Text:} william anders (born in british hong kong) was a crew member of nasa's apollo 8 alongside frank borman. william anders retired on september 1st, 1969. \\ \hdashline
    \multirow{2}{*}{\textbf{\agggen}} & \textbf{Sentence Planing: \textit{[birthPlace] [was\_a\_crew\_member\_of] [operator crewMembers] [backup\_pilot] [dateOfRetirement]}} \\
    & \textbf{Predicted Text:} william anders was born in british hong kong and served as a crew member on apollo 8. frank borman was a crewman aboard the nasa operated apollo 8 mission. the backup pilot was buzz aldrin. william anders retired on september 1st, 1969. \\ \hdashline
    \multirow{2}{*}{\textbf{{\agggen}$_{-\textup{AG}}$}} & \textbf{Sentence Planing: \textit{[birthPlace] [was\_a\_crew\_member\_of] [operator] [crewMembers] [backup\_pilot] [dateOfRetirement]}} \\
    & \textbf{Predicted Text:} william anders (born in british hong kong) was a crew member of apollo 8 operated by nasa, where buzz aldrin was a backup pilot. frank borman was also a crewman aboard apollo 8. buzz aldrin's backup pilot was buzz aldrin. william anders retired on september 1st, 1969. \\ \hdashline
    \multirow{2}{*}{\textbf{{\agggen}$_{-\textup{OD}}$}} & \textbf{Sentence Planing: \textit{[backup\_pilot] [was\_a\_crew\_member\_of] []birthPlace] [dateOfRetirement operator] [crewMembers]}} \\
    & \textbf{Predicted Text:} buzz aldrin was a crew member of apollo 8 with buzz aldrin as backup pilot and frank borman as buzz aldrin. william anders was a crewman aboard apollo 8. british hong kong born william anders retired. he was a backup pilot on the apollo 8 mission was operated by nasa. frank borman was also a crewman. \\ \hline \hline

    \multirow{7}{*}{\textbf{Input}} & school of business and social sciences at the aarhus university \textbf{\textit{[affiliation]}} european university association\\
    & school of business and social sciences at the aarhus university \textbf{\textit{[city]}} aarhus \\
    & school of business and social sciences at the aarhus university \textbf{\textit{[country]}} denmark \\
    & school of business and social sciences at the aarhus university \textbf{\textit{[dean]}} thomas pallesen \\
    & school of business and social sciences at the aarhus university \textbf{\textit{[established]}} 1928 \\
    & school of business and social sciences at the aarhus university \textbf{\textit{[latinname]}} universitas aarhusiensis \\ 
    & school of business and social sciences at the aarhus university \textbf{\textit{[numberofstudents]}} 16000 \\ \hdashline
    \textbf{Transf} & \textbf{Predicted Text:} the school of business and social sciences at the aarhus university in aarhus, denmark was established in 1928. it has 737 academic staff and 16,000 students. its dean is thomas pallesen and it is affiliated with the european university association. \\ \hdashline
    \multirow{2}{*}{\textbf{\agggen}} & \textbf{Sentence Planing: \textit{[country established city] [affiliation latinName numberOfStudents] [dean]}} \\
    & \textbf{Predicted Text:} the school of business and social sciences at the aarhus university in aarhus, denmark was established in 1928. it has 737 academic staff and 16,000 students. its latin name is "universitas aarhusiensis" and it is affiliated with the european university association. its dean is thomas pallesen. \\ \hdashline
    \multirow{2}{*}{\textbf{{\agggen}$_{-\textup{AG}}$}} & \textbf{Sentence Planing: \textit{[country] [established] [city] [affiliation] [latinName] [numberOfStudents] [dean]}} \\
    & \textbf{Predicted Text:} the school of business and social sciences at the aarhus university in denmark is affiliated with the european university association in brussels. the school was established in 1928 and is located in the city of aarhus. it is affiliated to the european univerrsity in the school the latin name of "universitas aarhusiensis". and there are 16000 students. its dean is thomas pallesen". \\ \hdashline
    \multirow{2}{*}{\textbf{{\agggen}$_{-\textup{OD}}$}} & \textbf{Sentence Planing: \textit{[latinName numberOfStudents] [country established] [dean affiliation] [city]}} \\
    & \textbf{Predicted Text:} the school of business and social sciences at the aarhus university in aarhus, denmark has 16,000 students and its latin name is "universitas aarhusiensis". it was established in 1928 in the country of denmark and it is affiliated with the european university association. the current dean is thomas pallesen and it has strong connections with the school also affiliated to its dean of business. and it also home to the school. \\ \hline \hline
  \end{tabular}
  \label{table:webnlg_res_extra}
  \vspace{-12pt}
\end{table}

\clearpage

\section{Factoid Sentence Segmentation}\label{sec:sup-Segmentation}

\begin{figure*}[!tbh]
\includegraphics[width=0.6\columnwidth]{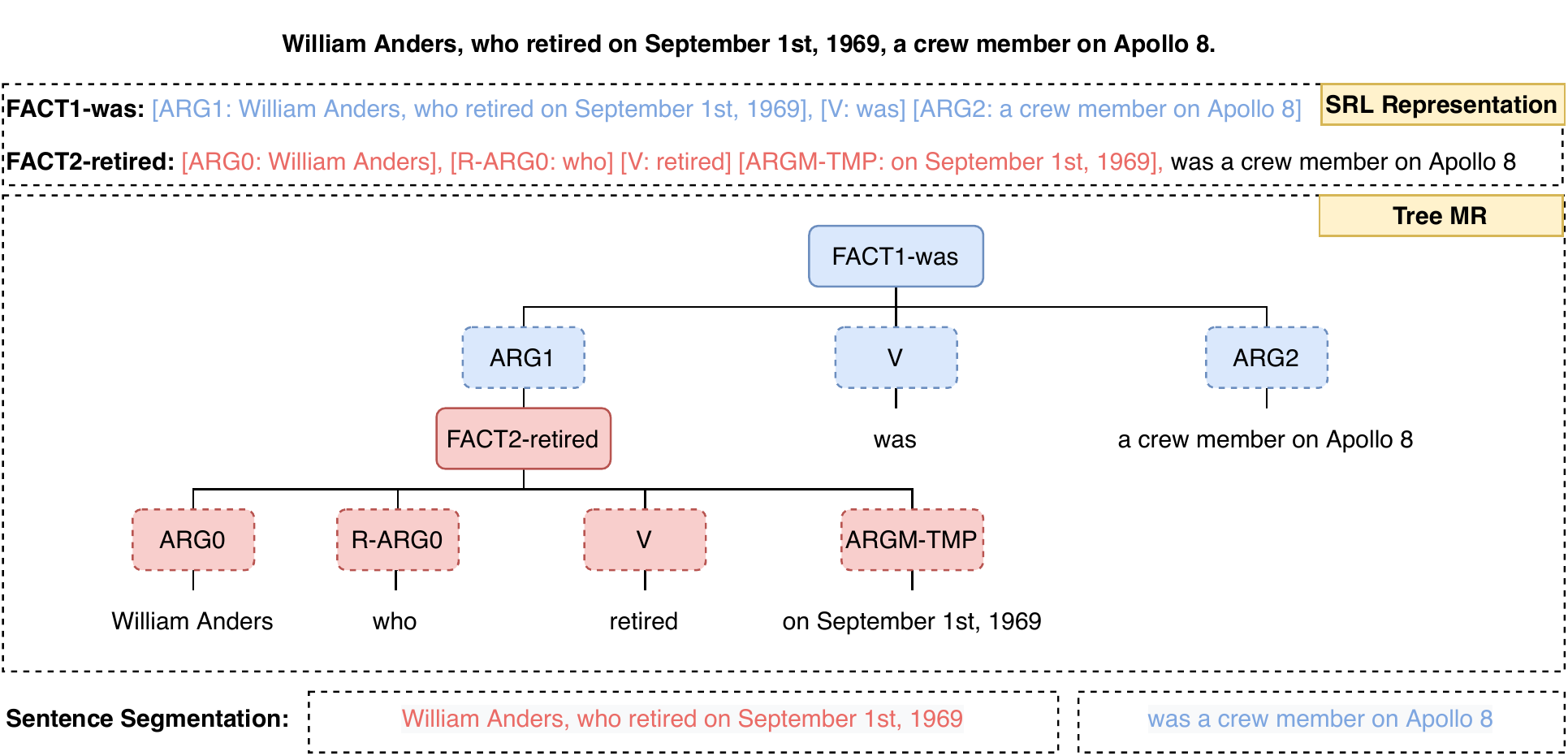}
\centering
\caption{Semantic Role Labeling based tree meaning representation and factoid sentence segmentation for text ``William Anders, who retired on September 1st, 1969, was a crew member on Apollo 8.''}
\label{fig:treemr}
\vspace{-12pt}
\end{figure*}

In order to align meaningful parts of the human-written target text to semantic triples,
we first segment the target sentences 
into sequences of facts using SRL, following \citet{xu2020fact}. 
The aim is to break down sentences into sub-sentences (\emph{facts}) that verbalize as few input triples as possible;
the original sentence can still be fully recovered by concatenating all its sub-sentences.  
Each \textit{fact} is represented by a
segment of the original text that  
roughly captures ``who did what to whom'' 
in one event.
We first parse the sentences into SRL propositions using the implementation of \citet{he2018jointly}.\footnote{The code can be found in \url{https://allennlp.org} with 86.49 test F1 on the Ontonotes 5.0 dataset.}
We consider each predicate-argument structure as a separate \textit{fact}, where the predicate stands for the event and its arguments are mapped to actors, recipients, time, place, etc.\ (see \autoref{fig:treemr}). The 
sentence segmentation consists of two consecutive steps: 

(1) {\em Tree Construction}, where we construct a hierarchical tree structure for all the facts of one sentence,  
by choosing the fact with the largest coverage as the root and recursively building sub-trees by replacing arguments with their corresponding sub-facts (ARG1 in FACT1 is replaced by FACT2). 

(2) {\em Argument Grouping}, where each predicate (FACT in tree) with its leaf-arguments corresponds to a sub-sentence. For example, in \autoref{fig:treemr}, leaf-argument ``was'' and ``a crew member on Apollo 8'' of FACT1 are grouped as one sub-sentence.

\section{Datasets}
\label{sec:sup-data-stats}

\paragraph{WebNLG.}
The corpus contains 21K instances (input-text pairs) from 9 different domains (e.g., astronauts, sports teams). 
The number of input triples ranges from 1 to 7, with an average of 2.9. 
The average number of \textit{facts} that each text contains is 2.4 (see Appendix~\ref{sec:sup-Segmentation}).
The corpus contains 272 distinct predicates. The vocabulary size for input and output side is 2.6K and 5K respectively.

\paragraph{E2E NLG.}
The corpus contains 50K instances from the restaurant domain.
We automatically convert the original attribute-value pairs to triples:
For each instance, we take the restaurant name as the subject and use it along with the remaining attribute-value pairs as corresponding predicates and objects.
The number of triples in each input ranges from 1 to 7 with an average of 4.4. 
The average number of \textit{facts} that each text contains is 2.6.
The corpus contains 9 distinct predicates. The vocabulary size for inputs and outputs is 120 and 2.4K respectively. 
We also tested our approach on an updated cleaned release 
\cite{dusek-etal-2019-semantic}.


\section{Hyperparameters}
\label{sec:sup-hyperparams}

\paragraph{WebNLG.} 
Both encoder and decoder are a 2-layer 4-head Transformer, with hidden dimension of 256. 
The size of token embeddings and predicate embeddings is 256 and 128, respectively. 
The Adam optimizer \cite{kingma2014ada} is used to update parameters. 
For both the baseline model and the pre-train of the HMM-based model, the learning rate is 0.1. 
During the training of the HMM-based model, the learning rate for the encoder-decoder fine-tuning and the training of the transition distributions is set as 0.002 and 0.01, respectively. 

\paragraph{E2E.}
Both encoder and decoder are a Transformer with hidden dimension of 128. 
The size of token embeddings and predicate embeddings is 128 and 32, respectively. 
The rest hyper-parameters are same with WebNLG.

\section{Parameterization: Emission Distribution}
\label{sec:sup-emission}
\begin{figure}[!tbh]
    \includegraphics[width=0.5\columnwidth]{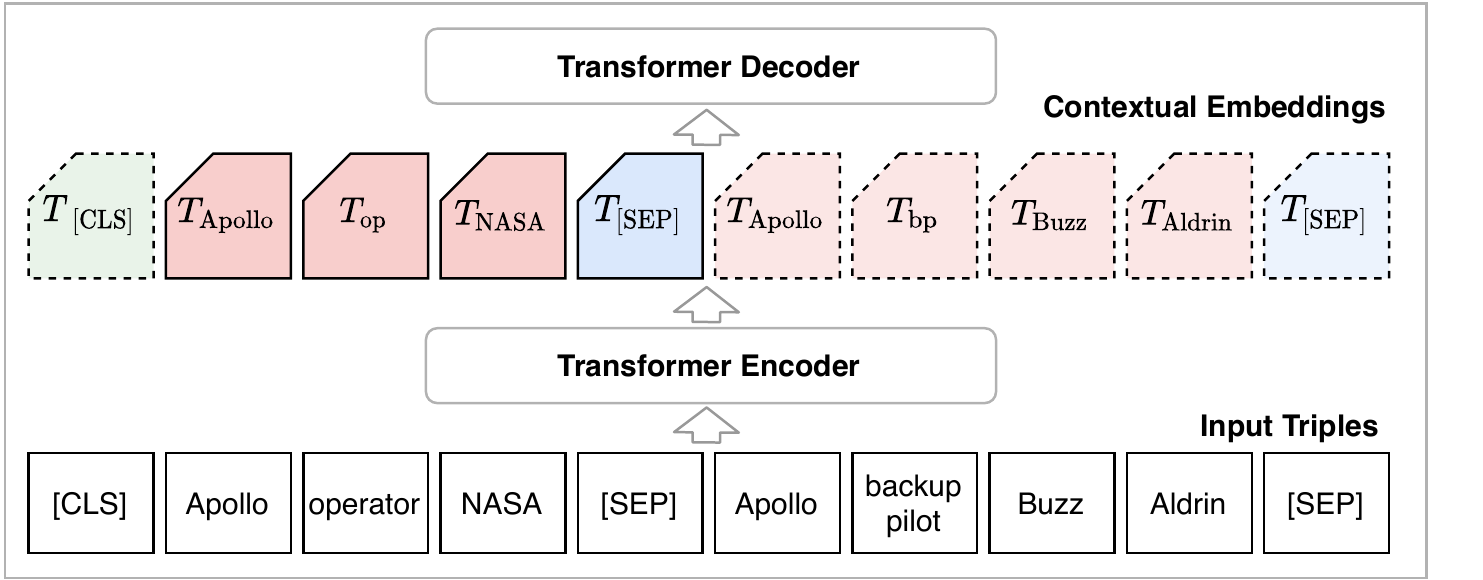}
    \centering
    \caption{
    The Transformer encoder takes linearized triples 
    and produces contextual embeddings 
    We assume that, at time step $t$, the Transformer decoder is generating \textit{fact} $\mathbf{y}_t$ conditioned on $\mathbf{z}_t$. The number of latent variables $L_t$ is 1. In other words, $\mathbf{z}_t=o_{t1}$. If the value of $o_{t1}$ is the predicate of the first triple (solid borders), then the second triple (dashed borders) is masked out for the encoder-decoder attention during decoding.}
    \label{fig:mask}
    \end{figure}

\section{Full Experiment Results on E2E}
\label{sec:sup-e2e_res}

\begin{table*}[!tbh]
  \small
  \centering
  \begin{tabular}{l|ccccccc|cccc}\hline
  {\bf Model} & {\bf Train} & {\bf Test} & {\bf BLEU} & {\bf NIST}  & {\bf MET} & {\bf R-L} & {\bf CIDer} & {\bf Add}  & {\bf Miss} & {\bf Wrong} & {\bf SER} \\ \hline\hline
  
  TGen$^\blacklozenge$ & \multirow{5}{*}{\rotatebox{90}{{\bf Original}}} & \multirow{5}{*}{\rotatebox{90}{{\bf Clean}}} & 39.23 & 6.0217 & 36.97 & {\bf55.52} & 1.7623 & {\bf00.40} & 03.59 & {\bf00.07} & {\bf04.05} \\ 
  Transformer &&& 38.57 & 5.7555 & 35.92 & 55.45 & 1.6676 & 02.13 & 05.71 & 00.51 & 08.35 \\
  \agggen &&& {\bf41.06} & {\bf6.2068} & {\bf37.91} & 55.13 & {\bf1.8443} & 02.04 & 03.38 & 00.64 & 06.06 \\ 
  {\agggen}$_{-\textup{OD}}$ &&& 38.24 & 5.9509 & 36.56 & 51.53 & 1.6525 & 02.94 & 03.67 & 02.18 & 08.80 \\
  {\agggen}$_{-\textup{AG}}$ &&& 30.44 & 4.6355 & 37.99 & 49.94 & 0.9359 & 08.71 & {\bf01.60} & 00.87 & 11.24 \\ \hline

  TGen$^\blacklozenge$ & \multirow{5}{*}{\rotatebox{90}{{\bf Clean}}} & \multirow{5}{*}{\rotatebox{90}{{\bf Clean}}} & {\bf40.73} & {\bf6.1711} & {\bf37.76} & {\bf56.09} & {\bf1.8518} & {\bf00.07} & {\bf00.72} & {\bf00.08} & {\bf00.87} \\ 
  Transformer &&& 38.62 & 6.0804 & 36.03 & 54.82 & 1.7544 & 03.15 & 04.56 & 01.32 & 09.02 \\
  \agggen &&& 39.88 & 6.1704 & 37.35 & 54.03 & 1.8193 & 01.10 & 01.85 & 01.25 & 04.21 \\ 
  {\agggen}$_{-\textup{OD}}$ &&& 38.28 & 6.0027 & 36.94 & 51.55 & 1.6397 & 01.74 & 02.74 & 00.62 & 05.11 \\
  {\agggen}$_{-\textup{AG}}$ &&& 26.92 & 4.2877 & 36.60 & 47.95 & 0.9205 & 05.99 & 01.54 & 02.31 & 09.98 \\\hline
  \end{tabular}
  \caption{Evaluation of \emph{Generation} (middle) and \emph{Factual correctness} (right) on the E2E NLG data (see Section~\ref{sec:Results} for metrics decription).}
  \label{table:e2e_res-sup}
  \vspace{-12pt}
  \end{table*}

\section{Annotation interface} \label{sec:sup-annotation}


\begin{figure*}[h]
  \centering
  \includegraphics[width=0.4\linewidth]{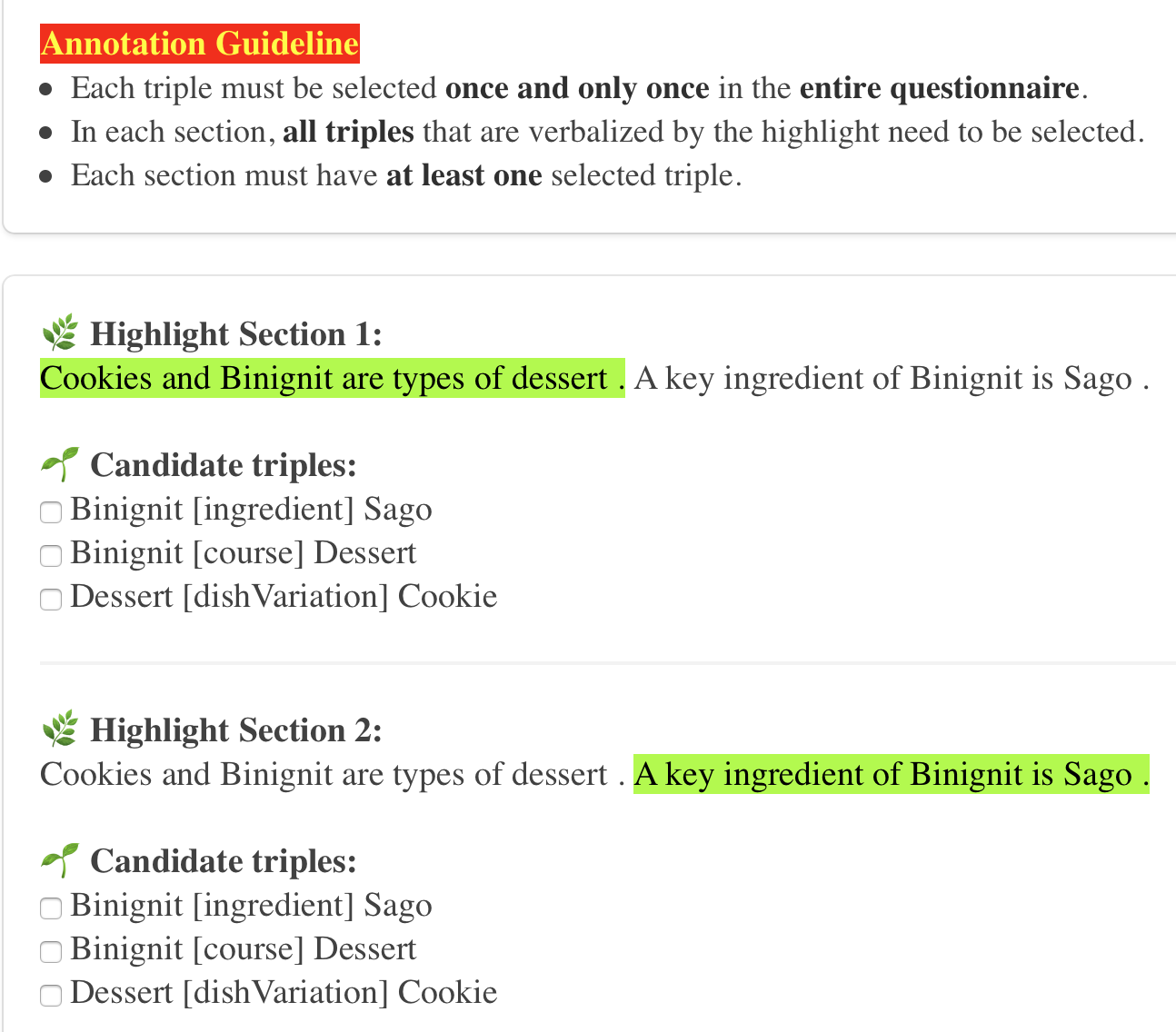}
  \caption{An example of the fact-triple alignment task (highlights correspond to facts).}
  \label{fig:HIT}
\end{figure*}

\end{document}